\documentclass[letterpaper, 10 pt, conference]{ieeeconf}  

\IEEEoverridecommandlockouts                              

\usepackage{graphicx}
\usepackage{xcolor}
\usepackage{amsmath} 
\usepackage{amssymb} 
\usepackage{array}
\usepackage{subcaption}
\usepackage[outdir=/fig/]{epstopdf}

\title{\LARGE \bf
 Human-inspired Grasping Strategies of Fresh Fruits and Vegetables Applied to Robotic Manipulation
}

\author{Romeo Orsolino$^{1*}$, Mykhaylo Marfeychuk$^{1*}$,  Mario Baggetta$^{2*}$, Mariana de Paula Assis Fonseca$^{1*}$, \\ Wesley Wimshurst$^{1}$, Francesco Porta$^{1}$, Morgan Clarke$^{1}$, Giovanni Berselli$^{2}$, and Jelizaveta Konstantinova$^{1}$
\thanks{This work was supported by the European Commission’s Horizon Europe Framework Programme under the IntelliMan project, GA 101070136.}
\thanks{*Authors contributed equally.}
\thanks{$^{1}$Ocado Technology, AL10 9UL, Hatfield, UK.
        \tt\small \{romeo.orsolino, mykhaylo.marfeychuck, mariana.fonseca, wesley.wimshurst, francesco.porta, morgan.clarke, j.konstantinova\}@ocado.com} %
\thanks{$^{2}$University of Genoa, 16126, Genoa, Italy.
        \tt\small \{mario.baggetta, giovanni.berselli\}@unige.it}}%

\begin{document}

\maketitle
\thispagestyle{empty}
\pagestyle{empty}

\begin{abstract}
Robotic manipulation of fresh fruits and vegetables, including the grasping of multiple loose items, has  a strong industrial need but it still is a challenging task for robotic manipulation.
This paper outlines the distinctive manipulation strategies used by humans to pick loose fruits and vegetables with the aim to better adopt them for robotic manipulation of diverse items. In this work we present a first version of a robotic setup designed to pick different single or multiple fresh items, featuring a multi-fingered compliant robotic gripper.
We analyse human grasping strategies from the perspective of industrial  key performance indicators (KPIs) used in the logistic sector.  The robotic system was validated using the same KPIs, as well as taking into account human performance and strategies. This paper lays the foundation for future development of the robotic demonstrator  for fresh fruit and vegetable intelligent manipulation, and outlines the need for generic approaches to handle the complexity of the task.

\end{abstract}

\section{Introduction}

\begin{figure}
	\centering
	\includegraphics[width=0.5\textwidth]{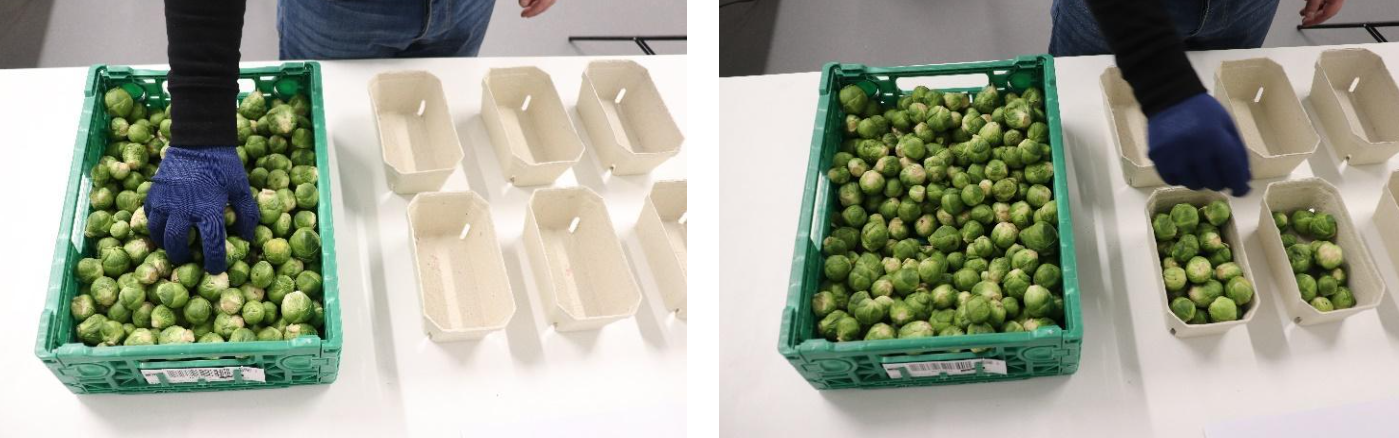}
	\includegraphics[width=0.5\textwidth]{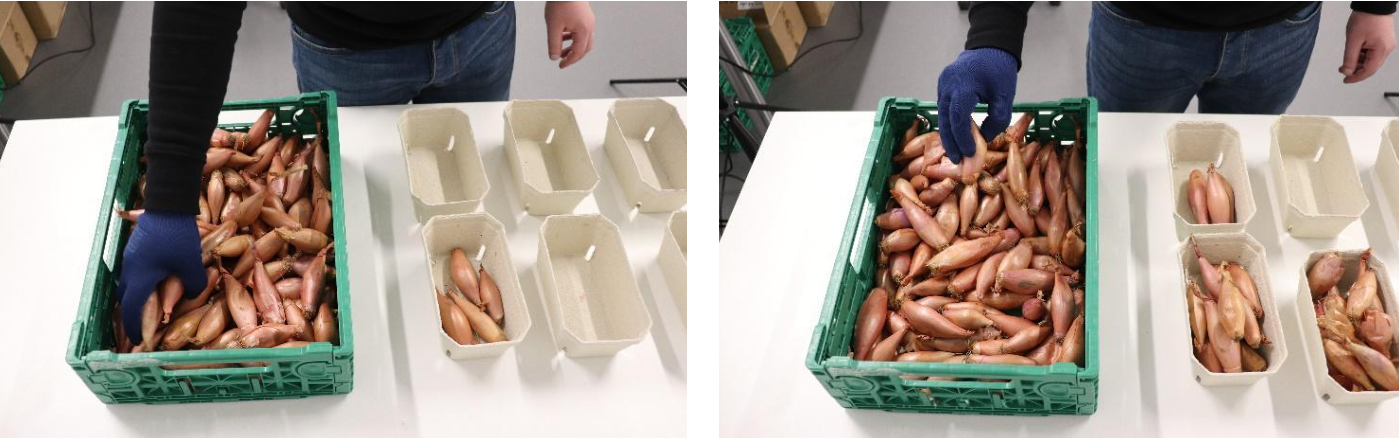}
	\includegraphics[width=0.5\textwidth]{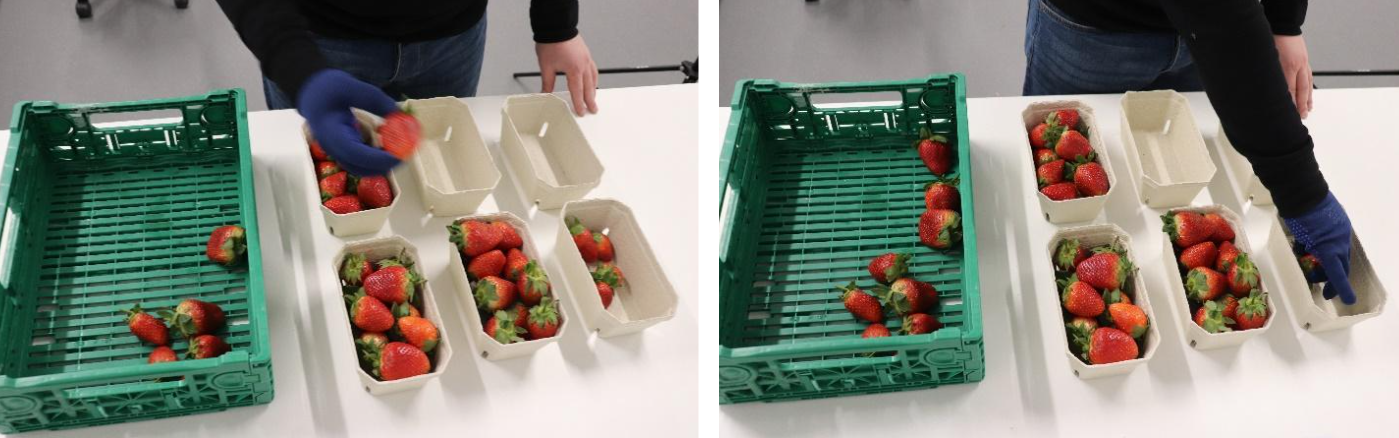}
	\caption{Samples of human grasp benchmark picking brussel sprounts (top), shallot onions (middle) and large strawberries (bottom). Our analysis included other fresh and loose products of small and large size such as sweet peppers, mandarins, button mushrooms, baby carrots and plum tomatoes.
	}
	\label{fig:human_grasp}
\end{figure}

Despite the great recent push from the academic community in robotic manipulation, the handling of multiple fresh and loose items remains a relatively unexplored topic. Most industrial implementations use robots for large scale picking of fruits and vegetables in mass production.
Such items present some particular challenges because of their irregular shape, delicate material and contamination risk.  In addition, it is important to recognise that the large variability of fresh fruits and vegetables that needs to be handled in a product distribution scenario (e.g., on-line shopping), significantly  reduces the feasibility of model-based approaches \cite{kleeberger2020, saxena2008}.

People are able to manipulate such items without difficulty, adapting their grasping strategies naturally. Studies have been carried out that provide useful insights on how humans pick individual rigid items \cite{Feix2016, grosse2016order} and on robotic multi-item grasping \cite{li2024grasp, chu2018multi, billard2019}. 
Little work has focused, however, on the robotic picking of multiple loose soft and delicate objects \cite{9812388, xu2023damage}.
Current approaches include using grippers with pneumatic suction membrane \cite{washio2022design}, compliant elements \cite{cheng2022reconfigurable, wang2022soft, baggetta2023design}, force sensors paired with machine learning algorithms \cite{karadaug2023non, mathew2021resoft}, and variable stiffness actuation systems \cite{friedl2020clash}. However, these solutions are primarily designed for single-object gripping, making multi-item grasping difficult.
Existing solutions specifically targeted towards multi-item grasping involve grippers with a large number of fingers \cite{makiyama2020pneumatic}, which can enclose many components at once but also easily damage them, and very large systems composed of multiple subsystems, each tailored to grasp a single object of predefined shape and dimension \cite{aoyama2022shell}.

EU IntelliMan project  \cite{IntelliMan} focuses on the question of “How a robot can efficiently learn to manipulate in a purposeful and highly performant way”. IntelliMan is developing a novel AI-Powered Manipulation System with persistent learning capabilities, able to perceive the main characteristics and features of its surroundings by means of a heterogeneous set of sensors, able to decide how to execute a task in an autonomous way and able to detect failures in the task execution in order to request new knowledge through the interaction with humans and the environment. As part of the IntelliMan project, there is a need to demonstrate robotic picking of various loose fresh fruits and vegetables  for logistic applications.


In this paper, we report the protocol and results of our benchmark of human grasping strategies that we carried out to tackle this challenge. In addition, we present a first version of a robotic setup composed of a robotic arm endowed with a custom multi-fingered gripper explicitly designed to enable the same grasping strategies observed on people.

This setup is used to train and validate new grasping policies for efficient and damage-free grasping in industrial use-cases \cite{Triantafyllou2019}. As a third contribution, we also propose the evaluation protocol and Key Performance Indicators (KPIs) used to benchmark custom grippers for industrial applications within the food sector.









\section{Human Demonstrations}

In our approach, we seek to understand the most efficient ways to manipulate objects in a robust way, and we believe that human demonstrations can provide invaluable insights for robotic picking enabling faster development cycles \cite{lin2013, leon2011, Cheng2018FastPL}.

\subsection{Methodology and Experimental Settings}


We carried out an experimental data collection with participants to study different grasping strategies used by people to pick and pack loose fruits and vegetables. For this, six participants were presented with a standardised storage crate containing six types of loose fruits or vegetables (one type at a time), where the goal is to pack fruits that are in good condition, into the punnets, as shown in Fig. \ref{fig:human_grasp}.

Participants performed two packing strategies for each type of object: 1) \textit{natural pick} - any comfortable picking strategy of their choice, using one hand, and 2) \textit{single pick},  each item is picked  individually using two fingered pinch grasp. Thus, we can outline a comparison between multi-fingered and two-fingered grippers.
The human hands and arm motion was recorded using Intel RealSense cameras, as well as  using visual markers tracked by an OptiTrack motion capture system. In total, over 200 data samples were collected, with each sample representing one punnet packing. Participants have performed from two to twelve grasping demonstrations to pack each  single punnet.

\subsection{Results from Human Grasping Experiments}



In the first part of the experiments, the participants were to employ a natural grasp. It was observed, that the natural grasping approaches could be classified into three distinct strategies:

\begin{itemize}
	\item \textit{Scooping with open wide fingers:} employs a scooping strategy with open hand exploiting all the benefits of the five fingers. This strategy was used for 50\% of the trials, generally preferred for smaller items.
	
	\item \textit{Scooping combined with a single item grasp:} uses a scooping strategy with the thumb opposing the palm. This strategy has been observed in 29.2\% of trials, and was observed to be used in conjunction with a single pick strategy.
	
	\item \textit{Multi pick:} employs multiple pinch grasps at a time, often picking from the different areas of the IFCO. This strategy has been observed in 20.8\% of the trials, and has been primarily used for larger item and for situations when the items are sparsely scattered in the container.
	
\end{itemize}

\begin{figure}[!t]
	\centering
	\includegraphics[width=\columnwidth]{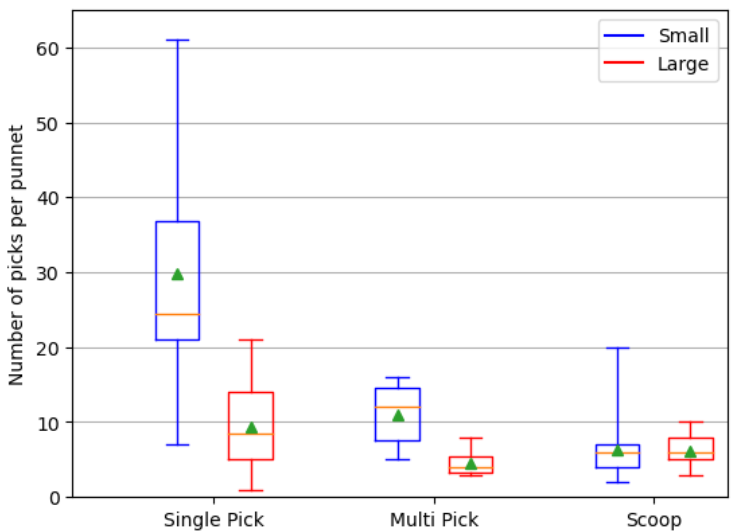}
	\caption{Number of picks and item type comparison for natural and single picks.}
	\label{fig:pick_strategy}
\end{figure}

In the second part of the experiments, the participants were asked to pick a single item at a time by employing a pinch grasp - \textit{thumb opposed to index and middle fingers together}. This strategy does not require any \textit{in-hand} rearrangement of the items and also implies a simpler packing to the target punnet (because of the single object), resulting in an overall faster motion.
However, the total Units (or \textit{punnets}) Per Hour (UPH) - as defined in Sec. \ref{sec:evaluation} - for the single pick are still lower compared to the natural pick, as shown in Fig. \ref{fig:pick_strategy}.

Comparing both of these experiments and taking into account the benchmarking methodology developed for robotic manipulation tasks \cite{Triantafyllou2019}, the following features were observed: 
\begin{itemize}
	\item Natural multi-fingered grasp leads to 70\% higher performance compared to a single item pinch grasp in terms of punnets packed - UPH. This grasp shows a 68\% reduction in the number of picks per punnet compared to a single item pinch grasp.

	\item People use a limited set of distinct strategies relying on the use of multiple finger manipulation to pick multiple loose items.

	\item During a multi-fingered natural grasp, people employ more complex packing and rearrangement strategies compared to a single grasp. This, however, does not impact the superior performance of multi-fingered grasp.

	\item The typical wrist rotation for all grasp types is 30° to 35° during the transport stage, which helps to distribute the applied force from fingers to the palm.
\end{itemize}

\section{Robotic Demonstrations}

The results of the experiments based on human grasping strategies have shown that humans naturally use three main gripping techniques. The most frequently used is scooping with open wide fingers, followed by scooping a single object using the opposable thumb, and multiple object gripping, which is particularly used for larger objects. These findings provide useful guidelines for designing the gripper used in the robotic setup developed by the University of Genoa.

\subsection{Multi-fingered Gripper Mechanical Design}\label{sec:gripper}

In particular, since the first two strategies require different finger positions, concentric for the first strategy and parallel for the second, the gripper must be capable of these configurations. Additionally, the third strategy requires each finger to be independently actuated so that it can grasp one object, reposition, and then grasp a second object. Finally, to ensure that smaller objects do not slip out of the grip, it is necessary to have at least four fingers to keep the opposing fingertips aligned during closure.

\begin{figure}[!t]
	\centering
	\includegraphics[width=\columnwidth]{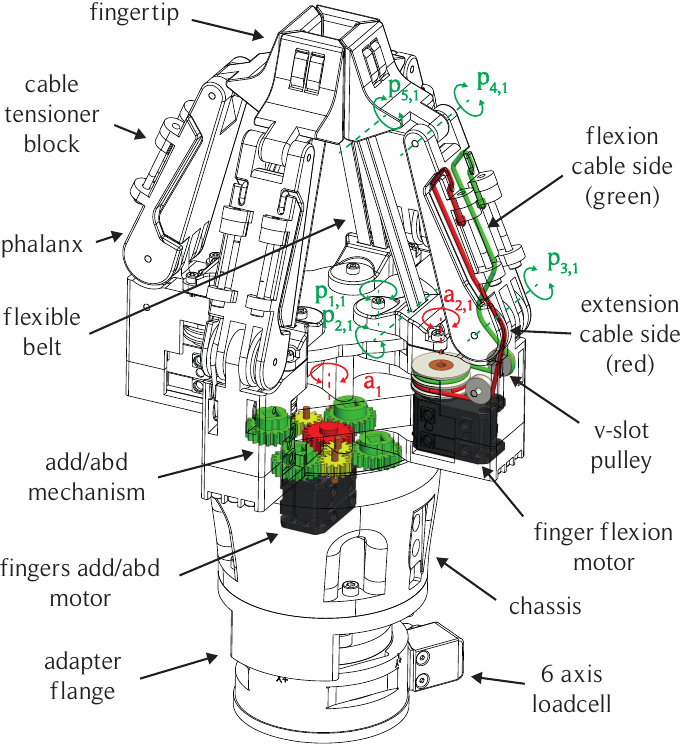}
	\caption{Mechanical design of the gripper, showing the kinematic joints of a single finger ($p_{1,n}$, $p_{2,n}$, $p_{3,n}$, $p_{4,n}$, $p_{5,n}$, with $n=1$ for the first finger, $n=2$ for the second, etc., in green), its degrees of actuation ($a_1$, $a_{2,n}$ for $n=1,2,3,4$, in red), the gear mechanism coupling the adduction degrees of freedom ($p_{1,n}$) of the four fingers, and the routing of the actuation cables.}
	\label{fig:gripper_design}
\end{figure}

Based on these requirements, the developed gripper, as shown in Fig. \ref{fig:gripper_design}, features a design with four fingers, each independently actuated by its own motor. Additionally, a fifth motor within the device’s chassis controls the adduction and abduction movements of all four fingers through a gear mechanism, enabling the transition between parallel and concentric configurations.

%
The flexion of the fingers is managed by a closed-loop cable system. This system comprises two cables: one dedicated to the flexion phase (depicted in green in the Fig. \ref{fig:gripper_design}) and another for the extension phase (shown in red). These cables are tensioned on one end by a screw/nut system situated on the phalanx of each finger, while the other end is securely connected to the motor's flange. The use of cable actuation ensures high precision in finger movements and also completely eliminates backlash, thereby enhancing the gripper's performance in delicate tasks.

\begin{figure}[!t]
	\centering
	\includegraphics[width=\columnwidth]{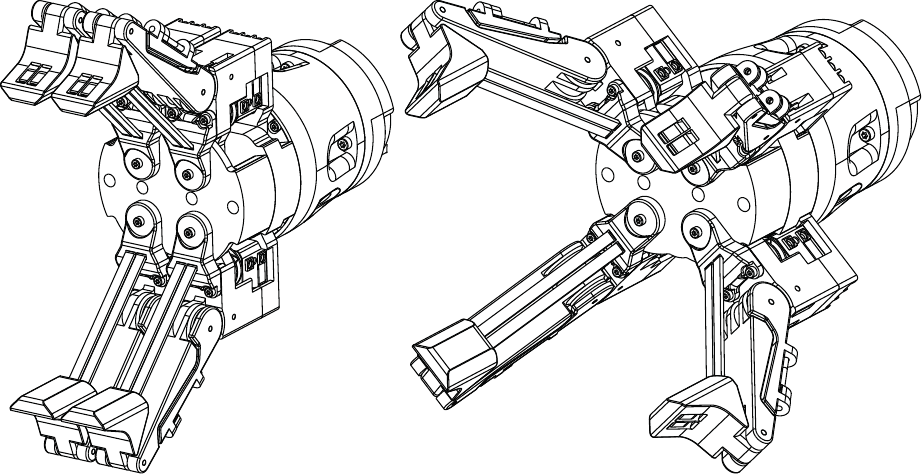}
	\caption{Possible gripping configurations of the gripper: parallel (left image) and spherical (right image). All intermediate positions between the two represented configurations are also possible.}
	\label{fig:grasp_pos}
\end{figure}

The actuation kinematics of the finger is based on the four-bar linkage principle, a well-established mechanism in mechanical engineering. In this design, one of the rigid links is replaced by a belt that exhibits rigidity in tension and flexibility in compression. The motors used for this device are Dynamixel XL-320 units, controlled by an OpenRB-150 board, which is also housed within the gripper's chassis. The design is completed with a connection flange that is compatible with coupling to a six-axis load cell, facilitating force and torque measurements. 
The integration of the fifth motor within the chassis allows the gripper to transition from a parallel gripping configuration to a spherical one. Additionally, it can achieve all intermediate positions between these two configurations (as shown in Fig. \ref{fig:grasp_pos}). This capability enables the gripper to grasp objects of various geometric shapes, such as cylindrical, spherical, and boxes.

\begin{figure}[!b]
	\centering
	\includegraphics[width=\columnwidth]{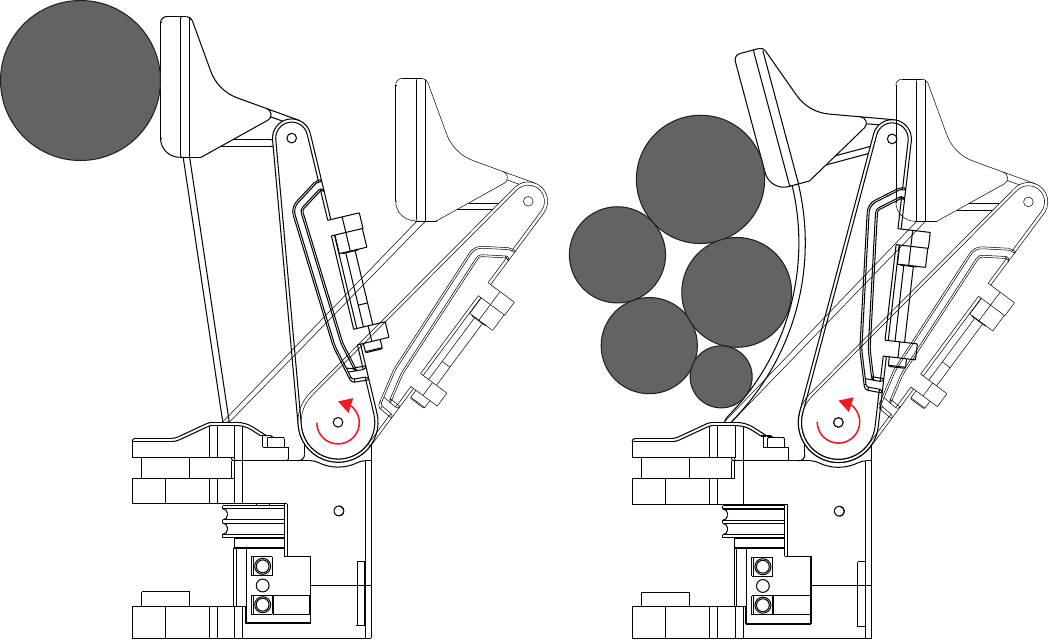}
	\caption{Behavior of the gripper's finger depending on the point of force application. If gripping using only the fingertip, the finger structure will behave like a rigid four-bar linkage (left image). If multiple objects are to be grasped, resulting in contact with the belt, the finger will exhibit flexible behavior and the fingertip will tend to close on the grip (right image).}
	\label{fig:finger_behaviour}
\end{figure}

The kinematic structure of the finger is designed to maintain a rigid and precise pinch grasp. At the same time, thanks to the use of the flexible belt, the gripper can adapt its behavior when grasping multiple objects or objects with unusual shape. In such scenarios, the finger transitions from a rigid parallel configuration to a flexible, adaptive one that conforms to the shape of the objects being grasped (as depicted in Fig. \ref{fig:finger_behaviour}). This adaptability enhances the gripper's utility and performance, making it highly effective in handling a diverse range of items with varying shapes and sizes.

\begin{figure}
	\centering
	\includegraphics[width=0.235\textwidth]{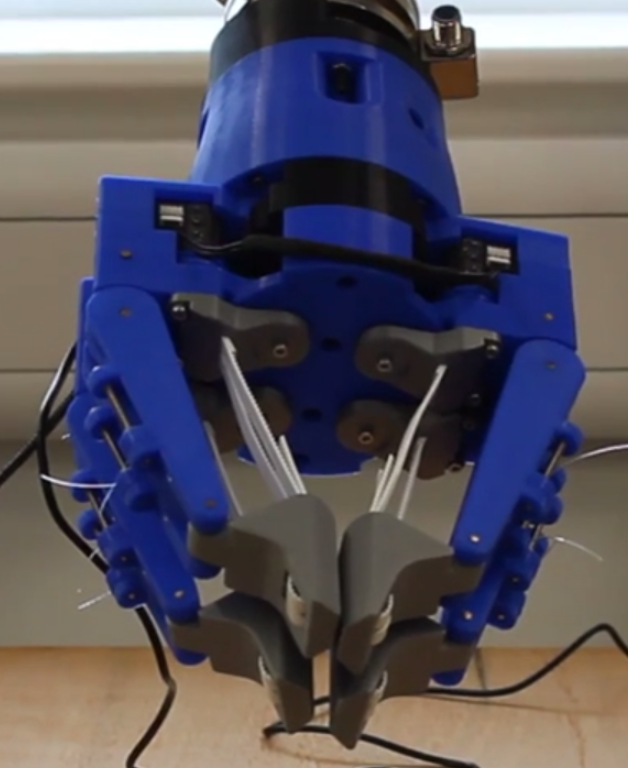}
	\includegraphics[width=0.235\textwidth]{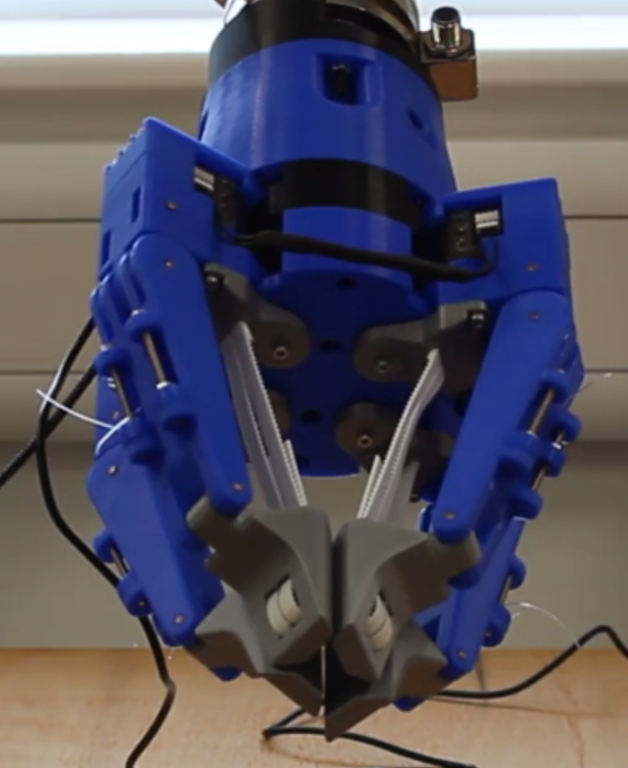}
	\caption{Gripper prototype in closed parallel (left) and concentric (right) configuration.}
	\label{fig:gripper_prototype}
\end{figure}

\subsection{Hardware Setup and Protocol}

Leveraging on the observations obtained from human demonstrations, we designed a robotic setup to implement fresh fruit and vegetable picking and packing, including the observed strategies and the custom multi-fingered gripper.
In this setup, we aim to evaluate the performance of the gripper in picking and placing single and multiple items. The same setup could be used to evaluate other grippers.

The robotic system is composed of a KUKA LBR iiwa 14 R820 robot manipulator equipped with an ATI Gamma Force/Torque sensor at its wrist, and the multi-fingered gripper developed by the University of Genova introduced in Sec. \ref{sec:gripper} (Fig. \ref{fig:gripper_design}). 

The robot is controlled by a task-space impedance controller and the desired trajectory is defined by a cartesian interpolation of start and end poses, with a pre-defined time duration. The stiffness of the controller was chosen as $\bold{K} = \mathrm{diag} \left(2000.0\bold{I}_3, 200.0\bold{I}_3\right)$, where $\bold{I}_3$ is a 3x3 identity matrix.
Furthermore, we used Behavior Tree \cite{Ghzouli2020} and Robot Operating System 2 (ROS2) \cite{Macenski2022} to move both gripper and robot arm.

\begin{figure}
	\centering
	\includegraphics[width=0.45\textwidth]{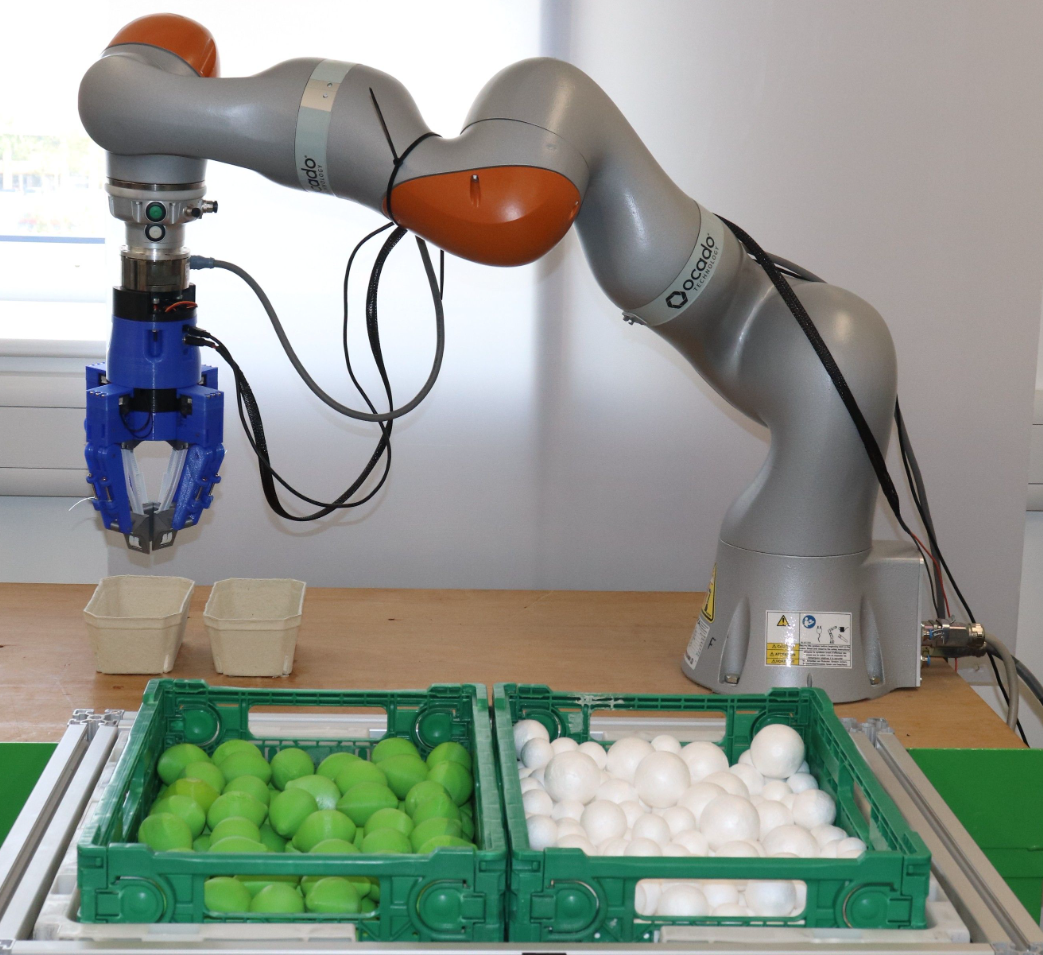}
	\caption{Experimental setup for testing robotic manipulation of soft and loose products using human-inspired grasping strategies. The robot is a KUKA LBR iiwa 14 robot equipped with a wrist ATI Gamma Force/Torque sensor and the multi-fingered  gripper developed by the University of Genova. Inside the two crates, the 3D printed limes (left) and polystyrene spheres (right) simulating fresh products can be found.}
	\label{fig:robot_setup}
\end{figure}

To evaluate the performance of the gripper we used two different objects as a substitute of the real fresh products: 3D printed \textit{limes} and polystyrene spheres (or \textit{pickles}) of different diameters between 5 and 50mm, illustrated in Fig \ref{fig:robot_setup}.
Two use cases were tested: picking a single item from a sparse crate, and picking multiple items at once from a full crate. In both cases, the robot was required to approach the picking area, grasp one or more items, transport the picked items and place them into a target punnet. Therefore, we defined five different phases:

\begin{itemize}
	\item \textit{Initialisation}: all the different modules (robot controller, motion planner, vision) are initialised and the configuration parameters are loaded.
	\item \textit{Approach}: the robot moves to a common starting point, opens the gripper (either in a parallel or concentric configuration), and then moves the gripper to a pose above the centre of the crate. This pose can have three different discrete \textit{approach angles} relative to the table, from 90° (vertical) to 60° with steps of 15°. We did not use lower approach angles than 60° as they would lead to collisions between the gripper and the crate and thus to potential damage. 
	\item \textit{Grasping}: the robot moves downwards towards the picking pose and closes the gripper as soon as a contact is detected by the wrist Force/Torque sensor.
	\item \textit{Transport}: the robot moves back to a pose above the picking pose, and then to the placement area.
	\item \textit{Placement}: the robot opens the gripper and drops the items into the target punnet.
\end{itemize}

In total, we had 24 different scenarios (2 objects x 2 picking types x 3 approach angles x 2 gripper configuration). For a more statistically significant evaluation, each scenario was repeated five times, totaling 120 runs.

The setup was identical for every trial, as illustrated in Fig. \ref{fig:robot_setup}. The robot is located on a table and a punnet is placed in front of it, indicating the placement area for the grasped items.
On the left side of the robot, there is a bench with two crates, where the robot picks the surrogate fruits/vegetables. The picking pose is pre-defined as a pose in the middle of the crate on the right. The same pose was used for both objects and both cases: single and multiple picking. The number of items the robot will pick depends on the amount of objects close to the pre-defined picking pose. 
In order to test if the robot was able to pick a single item, we set a few sparse items in the container and only one item in the picking location. To assess the multiple picking, the crate in the picking area was full of objects.

Since our goal is to evaluate the gripper, we pre-defined waypoints to each phase, and no recognition of object or picking/placing area were used.
In the grasping phase, the robot stops moving down when a given threshold on the magnitude of the force or torque measured by the wrist sensor is reached. The force threshold was set to 5N and the torque one to 1N.m. Moreover, a threshold on the derivative of the force and torque magnitude were also used (respectively 3N and 0.3N.m) to detect sudden impacts.

\subsection{Evaluation} \label{sec:evaluation}

To benchmark the gripper for different grasping strategies, we evaluate the following KPIs:

\begin{itemize}
	\item \textit{Time per Phase}: duration of each phase of our strategy (initialisation, approach, grasping, transport, and placement).
	\item \textit{Total Units Per Hour (UPH)}: number of products that can be picked and placed inside the target punnets in an hour.
	\item \textit{Success Rate}: percentage of successful trials over the total number of attempted picks.
\end{itemize}

\subsection{Results}
The evaluation of each KPI discussed in the previous Section is shown next:

\subsubsection{Time per Phase}
Table \ref{tab:time_per_phase} shows the average duration of each one of the five phases of the pick and placement task. 
Currently, the robot arm movements take a fixed amount of time, pre-defined by the user considering a maximum allowed velocity of the robot which is low for safety reasons.
The Initialisation phase has a robot movement duration of 5 seconds. The Approach and Transport phases currently has trajectories of 10s each, but this value could be further reduced.
The Grasping phase duration is determined by the speed of downward motion of the gripper, the sensitivity of the wrist F/T sensor and by the amount of objects in the crate.
The 210ms for the Placement duration corresponds to the fingers opening time and is lower limited by the maximum speed of the fingers.

\begin{table}
\begin{center}
\caption{Average phase duration}
\label{tab:time_per_phase}
\begin{tabular}{ c c c c c c c }
	\hline
	$\bold{Phase}$ & Init. & Appr. & Grasp. & Trans. & Place. & $\bold{Total}$ \\
	\hline
	$\bold{Time [s]}$ & 5.12 & 10.37 & 5.80 & 20.12 & 0.21 & $\bold{41.68}$ \\
	\hline
\end{tabular}
\end{center}
\end{table}

\subsubsection{UPH}
During the benchmarking execution, we paused the system between each trial. The total experiment time used to calculate the UPH does not include the downtime between the trials, but the total time spent when the experiments were running, for both success and failure cases. With that in mind, we obtained 47.93 successful Trials Per Hour (TPH), as shown in Table \ref{tab:uph}.
This value does not take into account how many items were picked in each trial. 
For the multiple item case, the system was able to successfully pick and pack between one and four items. Taking into account all the successful packed items, totaling 56, we obtained a UPH of 65.5.
In particular, our benchmark scored a UPH of 57.1 for the single-item scenario and a UPH of 74.2 for the multi-item scenario. This confirms the observation from our study on the human grasping strategies, that multi-item picking results in superior performance despite the incresed difficulty of the task.

\begin{table}
	\begin{center}
		\caption{UPH per item picked and placed}
		\label{tab:uph}
		\begin{tabular}{ c c }
			\hline
			$\bold{Description}$ & $\bold{Value}$  \\
			\hline
			Total trials & 120 \\
			Successful trials & 41	\\		
			Placed items & 56 \\						
			Total time [s] & 3079.76 \\
			TPH & 47.9 \\
			\textbf{UPH} & $\bold{65.5}$ \\
			\hline
		\end{tabular}
	\end{center}
\end{table}

\subsubsection{Success rate}
\begin{figure}
	\centering
	\includegraphics[width=0.5\textwidth]{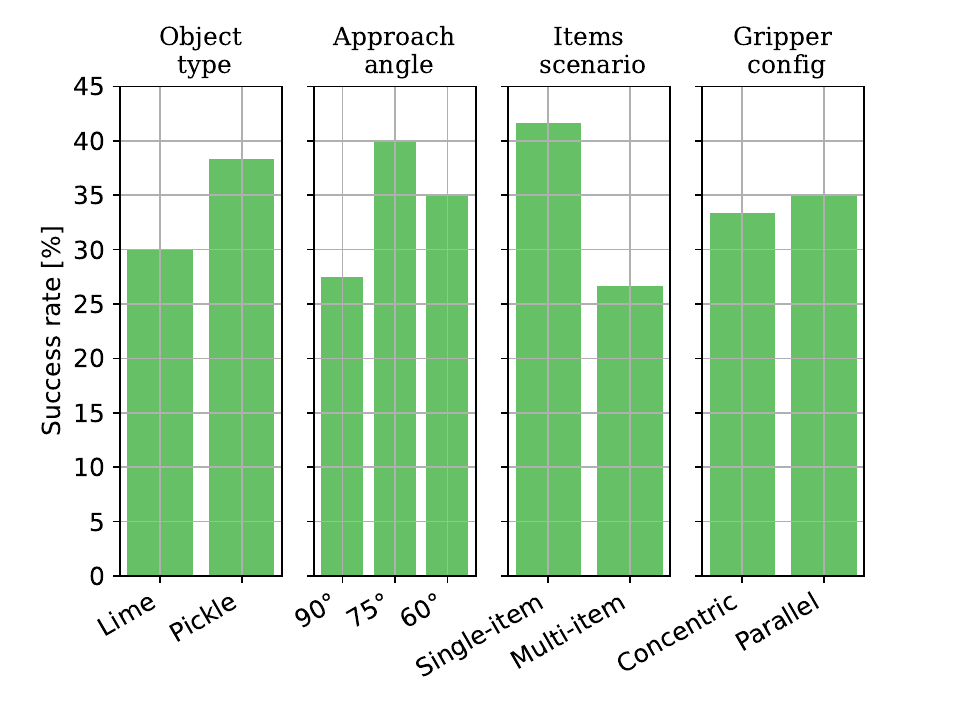}
	\caption{Picking success rate (from left to right) per object type, approach angle, pick type, and per gripper configuration.}
	\label{fig:success_rate}
\end{figure}

As our goal is to analyse the performance of the gripper, we ignore here the system failures that are not due to the gripper or to the grasping strategy. We had 41 successful runs out of the 63 trials that reached the grasping phase, leading to a success rate of $65.08\%$. Figure \ref{fig:success_rate} shows the success rate per object type (top left), approach angle (top right), pick type (bottom left), and gripper configuration (bottom right). 
We can see no significant difference between the gripper configurations (bottom right) while there were larger differences in success rate for different object and picking types.

To better understand the success cases, we also plot the number of successful trials  per object divided by pick type (Fig. \ref{fig:success_and_failure}, left). 
We can see that the single lime scenario obtained the highest success rate, and the multiple limes, the lowest. This reflects the difficulty of the gripper to penetrate its fingers among the rigid 3D printed limes. Such challenge could be overcome by an adaptive strategy for objects of different compliance, which is left to our future work.

\subsection*{Reasons for Failure}
\begin{figure}
	\centering
	\includegraphics[width=0.5\textwidth]{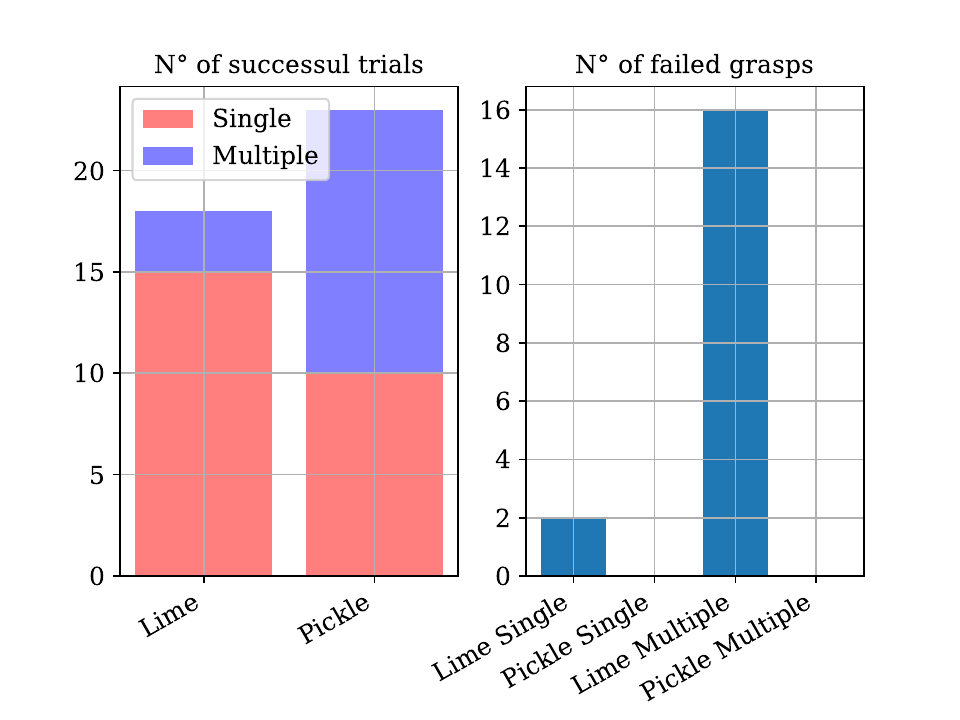}
	\caption{Success trials per object type and pick type (left) and grasp failures per scenario (right).}
	\label{fig:success_and_failure}
\end{figure}

Excluding the system/engineering failures and focusing on the benchmark process failures, we mainly encountered two issues: grasping and dropping failures. By grasping failure we mean that the robot reached the picking phase and approached the object but was not able to grasp it. The dropping failure mode, instead, consists of items being dropped outside of the placement punnet. The grasping and dropping failures were responsible for, respectively, 82\% and 18\% of the observed process failures. Furthermore, no grasping failure was observed while tackling soft items (in single- or multi-item scenarios), with all of the grasping failures happening while trying to pick the harder 3D printed limes (see Fig. \ref{fig:success_and_failure}, right). This was due to the difficulty of the fingers to penatrate among the items sufficiently to establish a stable grasp with the items. 

\section{Conclusion}

Grasping multiple loose items, as done by humans during packing, is a challenging task for robotic platforms that requires appropriate choice of motion planning strategies and gripper design. In this paper, we presented the study we conducted on human participants to understand their preferred grasping strategies to pick and place large quantities of fresh and loose products. We outlined distinctive human grasping strategies that provide a useful input to the design of robotic hardware and software setups. We presented a new custom tendon-driven multi-fingered gripper prototype which was designed based on the insights from our human grasping analysis and we described the hardware setup and protocol that lays the foundations for a new robotic demonstrator. By implementing human-inspired strategies on the proposed robot setup we have demonstrated the observation from our human benchmarking campaign, that multi-item pick leads to superior pick and pack performance compared to single-pick strategy, despite the increased complexity of the task.

We reported the details of our experimental protocol which can be replicated to benchmark the performance of new robotic grippers and grasping policies for the pick and placement task of multiple loose products.
The further development of the robotic demonstrator is ongoing, and it is planned to exploit machine learning applications, as well as leverage on iterative hardware design improvements to develop more advanced grippers.


\addtolength{\textheight}{-16cm}   





\bibliographystyle{IEEEtran}
\bibliography{mybibfile}

\end{document}